\documentclass[a4paper]{./styles/svproc}
\usepackage[backend=biber,
            url=false,
            isbn=false,
            doi=false,
            backref=false,
            natbib=true,%compatibility aliases
            mincitenames=1,
            maxcitenames=1,
            citestyle=numeric-comp,
            sorting=nyt,%none, 
            block=none]{biblatex}
\usepackage{url}

\usepackage[bookmarks=true]{hyperref}
\usepackage{graphicx}
\graphicspath{{figures/}}
\usepackage{multirow}
\usepackage{color}
\usepackage{caption}
\usepackage{booktabs}
\usepackage{amsmath}
\usepackage{amssymb}
\usepackage{bm}
\usepackage{verbatim}
\usepackage{gensymb}
\usepackage{threeparttable}
\usepackage{enumitem}
\usepackage{comment}
\usepackage{xcolor}
\usepackage{xspace}
\usepackage{wrapfig}
\usepackage{siunitx}
\usepackage{booktabs}
\usepackage{array}
\usepackage{makecell}
\newcommand{\ba}{\mathbf{a}}

\newcommand{\bs}{\mathbf{s}}

\renewcommand{\remark}[3]{{\color{#2}[#1: #3]}}
\newcommand{\daniel}[1]{\remark{Daniel}{cyan}{#1}}

\newcommand{\ie}{i.e.,\xspace}
\newcommand{\eg}{e.g.,\xspace}

\newcommand{\method}{SOPE\xspace}

\begin{document}
\mainmatter              % start of a contribution
\title{Learning to Singulate Objects in Packed Environments using a Dexterous Hand}
\titlerunning{Singulating Objects in Packed Environments}  % abbreviated title (for running head)
\author{Hao Jiang\inst{1} \and Yuhai Wang\inst{1}$^{,*}$ \and Hanyang Zhou\inst{1}$^{,*}$ \and Daniel Seita\inst{1} \\{\footnotesize * equal contribution}
}
\authorrunning{Jiang et al.} % abbreviated author list (for running head)
\tocauthor{Hao Jiang, Yuhai Wang, Hanyang Zhou, Daniel Seita}
\institute{University of Southern California, Los Angeles, CA 90089, USA,\\
\email{\{hjiang86,yuhaiwan,zhouhany,seita\}@usc.edu} %,\\ WWW home page: \texttt{http://users/\homedir iekeland/web/welcome.html} % maybe we put our SOPE website here
}

\maketitle              % typeset the title of the contribution

\vspace{-1em}
% Daniel: for our method's results, I am using (40 + 31) / (45 + 40) to determine this. 
\begin{abstract}
Robotic object singulation, where a robot must isolate, grasp, and retrieve a target object in a cluttered environment, is a fundamental challenge in robotic manipulation. 
This task is difficult due to occlusions and how other objects act as obstacles for manipulation. A robot must also reason about the effect of object-object interactions as it tries to singulate the target.
Prior work has explored object singulation in scenarios where there is enough free space to perform relatively long pushes to separate objects, in contrast to when space is tight and objects have little separation from each other. 
In this paper, we propose the Singulating Objects in Packed Environments (\method) framework. 
We propose a novel method that involves a displacement-based state representation and a multi-phase reinforcement learning procedure that enables singulation using the 16-DOF Allegro Hand. 
We demonstrate extensive experiments in Isaac Gym simulation, showing the ability of our system to singulate a target object in clutter. We directly transfer the policy trained in simulation to the real world. Over 250 physical robot manipulation trials, our method obtains success rates of 79.2\%, outperforming alternative learning and non-learning methods. 
Supplementary material is available at \url{https://sope-dex.github.io/}. 
% The abstract should summarize the contents of the paper using at least 70 and at most 150 words. It will be set in 9-point font size and be inset 1.0 cm from the right and left margins. There will be two blank lines before and after the Abstract. \dots We would like to encourage you to list your keywords within the abstract section using the \keywords{...} command.
\end{abstract}
 
\section{Introduction}
 
A fundamental challenge in robotic manipulation is singulating a target object in ``packed'' environments that have multiple objects in clutter \emph{and} limited space for a robot to maneuver and adjust the objects.  
This is common in daily life; examples include retrieving a book from a bookshelf, pulling out items from a stuffed bag, or retrieving a subset of clothing from a stack of clothes. 
In contrast to when a robot grasps objects isolated in free space, in cluttered environments with multiple objects, a robot must reason about challenges ranging from visual occlusions to the complex interactions between objects as they make contact with each other. 
Furthermore, in tightly-constrained environments which limit the extent to which objects can move, it can be difficult to singulate the target if other objects block or hinder the movement of a robot's gripper. 

% Daniel: only if we were trying to minimize environmental perturbations.
%if objects have little separation from each other, a robot that retrieves an object may also need to  minimally perturb the ``distractor'' objects. For example, with extracting an book from a crowded bookshelf, it would be desirable if a robot could retrieve the target without knocking over all the other books. 

We propose to formalize these problems under the \emph{\textbf{S}ingulating \textbf{O}bjects in \textbf{P}acked \textbf{E}nvironments \textbf{(SOPE)}} framework, where a robot must singulate a target item in a setting which makes it difficult to fully isolate the target. 
By ``singulating,'' we refer to the complete procedure where a robot \emph{isolates} the target, then \emph{grasps} and \emph{retrieves} it. 
As a concrete task in the SOPE framework, we consider a container which holds several rigid removable blocks, and the robot is given one target block to singulate. 
The robot must ultimately retrieve only the target block. 
%See Figure~\ref{fig:teaser} \daniel{might change} for examples of our system deployed in the real world under diverse singulation scenarios with different targets. 

\begin{figure}[t]
\centering
\includegraphics[width=1.00\textwidth]{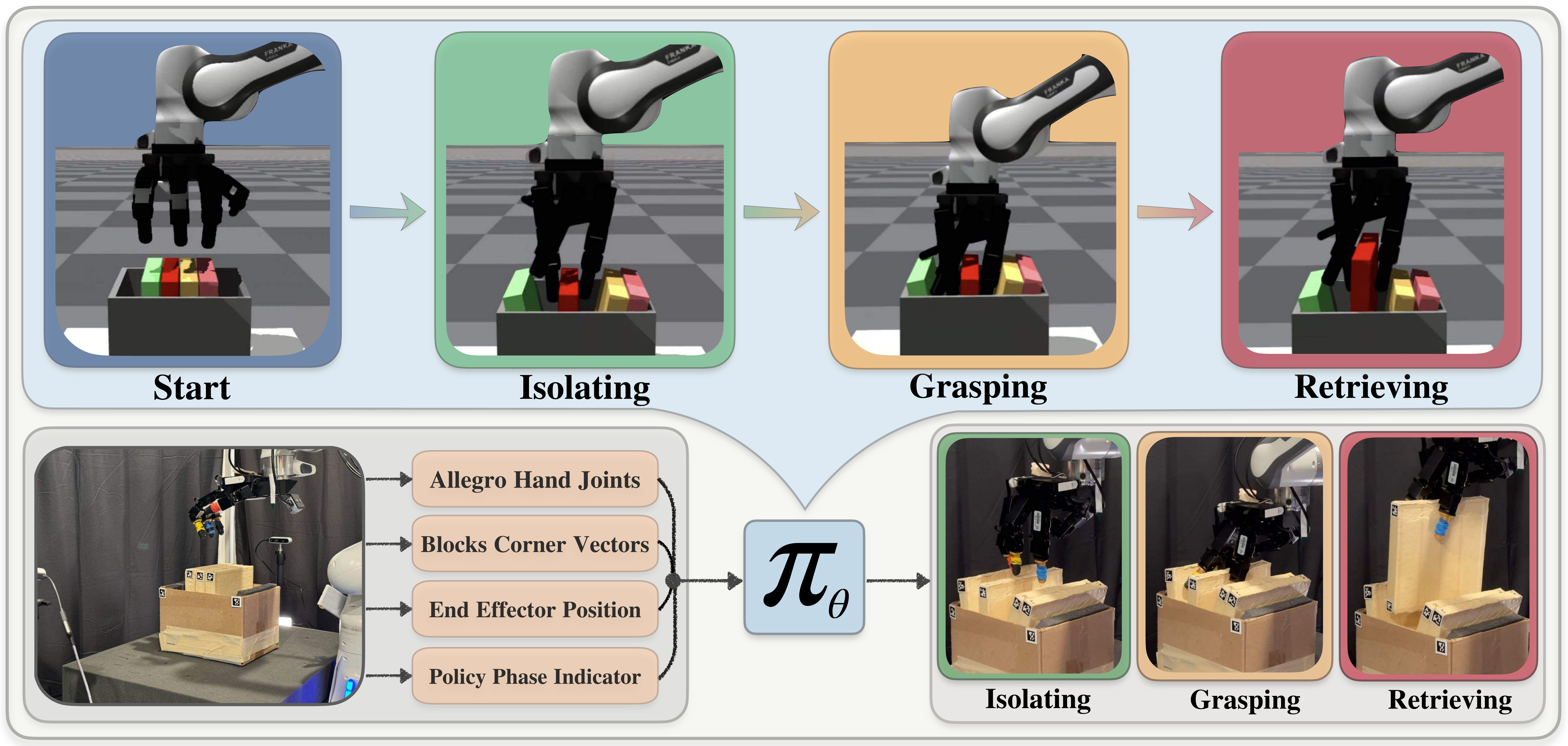}
\caption{
  Overview of our proposed \method framework for object singulation, shown here with the Allegro Hand. The task involves three phases: isolating, grasping, and retrieving. We train a policy in simulation using phase-dependent reward functions. The policy uses the hand joints, the block positions, and the robot arm in the state representation. We directly transfer the trained policy to the real world. 
}
\vspace{-15pt}
\label{fig:overview_of_system}
\end{figure}

In this work, we use deep reinforcement learning to train a singulation policy using the 16-DOF Allegro Hand in simulation. 
This dexterous hand enables the robot to extract the target while simultaneously stabilizing or pushing aside the other items. 
We design a novel multi-phase reinforcement learning procedure with a displacement-based state representation and a phase-dependent reward function which consider the ``distance'' between a target object and its nearby objects; a dexterous manipulator is trained to maximize this distance. 
We directly transfer the policy to the real world, where the policy can singulate one of multiple possible target blocks. 
See Figure~\ref{fig:overview_of_system} for an overview of our pipeline.

To summarize, the contributions of this paper include: 

\begin{enumerate}[noitemsep,leftmargin=*]
    \item A novel simulation-to-real pipeline for a dexterous object singulation system which includes a state representation and reward function that both consider the distance between the target and neighboring objects. 
    \item A simulation environment for training and benchmarking object singulation. 
    \item \textbf{250 physical experiments} suggesting that our method obtains 79.2\% success rates when singulating one item out of three or one item out of four, outperforming alternative learning and non-learning methods. 
\end{enumerate}

\section{Related Work}

\subsection{Object Singulation and Manipulation in Packed Environments}

Object singulation, where the objective is to isolate, grasp, and retrieve an object from clutter, is a fundamental problem in robotics. Early work on singulation has studied algorithmic pile interaction strategies~\cite{Interactive_singulation_2012}. Later works incorporate machine learning for singulation to determine pushing actions~\cite{singulate_push_2017,split_DeepQ_singulation_2020}. 
For example,~\cite{zeng2018learning} show how to synergize pushing and grasping, to enable retrieval of (initially) ungraspable objects. 
However, these preceding works consider tabletop singulation scenarios with substantially more free space than what we consider, which involves items packed together in a setting that limits pushing ranges.

Prior work in object singulation has also considered deformable objects. For example,~\cite{tirumala2022reskin} study singulating layers of cloth, and~\cite{slipbagging2023,autobag2023} show how to separate thin layers of plastic bags. 
These methods require heavy engineering of the physical setup or careful design of the action primitives (\eg ``shaking'' or ``pressing'') whereas we focus on using dexterous manipulators without using action primitives.  In addition, these works use parallel-jaw grippers, and we use higher-DOF grippers for increased flexibility.  
In independent and concurrent work,~\cite{inhand_singulation_2024} study in-hand singulation with a custom 5-DOF gripper and tactile sensing.

% Daniel: removed \emph{stowing}~\cite{chen2023stowing} to save space. 
Finally, other research has studied tasks such as insertion in packed environments with tight tolerance~\cite{li2022seehearfeel,ai2024robopack}. In some sense, this is the inverse of singulation, as we focus on \emph{retrieving} an object in clutter. These works also primarily focus on combining vision and touch and use large tactile sensors with a form factor that can inhibit singulation in a packed environment. 
%(\eg GelSight~\cite{yuan_gelsight_2017}) 

\subsection{Dexterous Manipulation}
 
%By dexterous manipulation, our focus here is on single-arm manipulation using a high-DOF gripper such as the Shadow Hand or Allegro Hand. 
Dexterous manipulation has been studied for decades in robotics~\cite{mason1985robot,salisbury1982articulated}. Classical works include~\cite{bai2024palm,kumar2024inhand} which focus on analytical model-based trajectory optimization. While effective, these models can be difficult to use for more complex and diverse manipulation. 
Consequently, researchers have turned to approaches which either use pre-defined skills for in-hand manipulation~\cite{bhatt2021robustinhand} or which use machine learning. 
Among the latter, prior work has shown dexterous manipulators capable of adjusting a Rubik's cube to reach a solved configuration~\cite{rubik_cube_2019}, reorienting with the hand facing downwards~\cite{chen2023visual}, rotating without visual information~\cite{touch-dexterity}, and grasping objects based on their functions~\cite{agarwal2023dexterous}. 
%In contrast, we are interested in dexterous object \emph{singulation}. 
Other research focuses on tasks such as vegetable peeling~\cite{chen2024vegetables}, pen spinning~\cite{wang2024penspin}, untwisting lids~\cite{lin2024twisting}, or playing piano in simulation~\cite{robopianist2023}. 
These works specialize to different problem settings. In contrast, we focus on single-arm dexterous singulation among multiple items. %and we do not need a second manipulator. 
%in-hand translation~\cite{yin2024learninginhandtranslation}

%HATO~\cite{lin2024learning}, 
%, OPEN TEACH~\cite{iyer2024open},
In this work, we use ``dexterous'' to refer to manipulation using high-DOF grippers, though there are still a range of complex tasks which a robot can perform with ``simple'' parallel-jaw grippers. 
%Nonetheless, there are still a wide range of dexterous manipulation tasks which a robot can perform with a ``simple'' gripper~\cite{zhou2022ungraspable}.
As data is often a bottleneck, recent work has designed low-cost teleoperation systems for \emph{bimanual} dexterous manipulation with parallel-jaw grippers such as ALOHA~\cite{Zhao-RSS-23} and GELLO~\cite{wu2023gello}. 
Researchers have also built teleoperation systems for higher-DOF grippers, of which recent examples include DexCap~\cite{wang2024dexcap} and Bunny-VisionPro~\cite{bunny_visionpro}. 
These teleoperation systems have great potential to accelerate real world data collection, which can be utilized by imitation learning algorithms~\cite{chi2023diffusionpolicy} to learn complex tasks. 
These works are orthogonal to our approach, which focuses on zero-shot simulation-to-real transfer using a single dexterous gripper trained with reinforcement learning, without teleoperation. % system, though we will consider using teleopration in future work. 

% https://twitter.com/LerrelPinto/status/1638563384156200962
Recent work has also explored sample-efficient imitation learning approaches with application to dexterous tasks that involve object singulation~\cite{SeeToTouch2024,guzey2023dexterity}. These works rely on expensive tactile sensors, whereas we do not use tactile sensors. 
In this paper, we also use reinforcement learning. Prior work has shown real world learning of dexterous manipulation for adjusting beads and rotating valves~\cite{zhu2020realworldRL}. While effective, this may struggle with finer-grained object singulation tasks. %and requires real-world learning. 
We use the Allegro Hand in this work, which is too delicate to use with real-world reinforcement learning and hence requires simulation; we will consider other hands such as LEAP~\cite{shaw2023leaphand} in future work. 
%We will also explore complementary techniques such as masked visual pre-training~\cite{radosavovic2022pretraining} which has been used with dexterous manipulation.  
Among the most relevant prior work includes~\cite{chen2023sequential} which develops a procedure to chain different policies for obtaining and then stacking lego blocks using the Allegro Hand. In contrast, we focus on singulation, and use an alternative multi-phase RL procedure, coupled with an efficient state representation that aids singulation. 
\vspace{-3pt}

\section{Problem Statement}
\vspace{-3pt}

% Daniel (July 21): I polished up this section, what do you think?

We assume a single-arm robot with a high-DOF, multi-fingered dexterous manipulator. 
In front of the robot is a flat workspace with a box that contains $n$ blocks $\mathcal{B} = \{b_1, \ldots, b_n\}$. %in this paper, we test with $n \in \{3,4\}$. 
The robot is given an integer $i \in \{1, 2, \ldots, n\}$ which indicates the index of the target block, and must lift block $b_i$ above a height threshold, \emph{without} lifting any other block. 
We assume that the robot can accurately estimate the keypoints describing the state of the blocks. The collection of keypoints forms part of the environment state $\bs_t$ at time $t$. 
The robot's action $\ba$ at time $t$, \ie $\ba_t$, consists of the change in the joint angles of the hand only (not the robot arm).  
We frame this problem as determining a policy $\pi_\theta$ parameterized by $\theta$ which produces actions from states at each time step: $\pi_\theta(\bs_t) = \ba_t$.  
A \emph{trial} is an instance of this object singulation task, which starts when the robot hand is above the set of blocks, and terminates upon either task success, or reaching a time limit without succeeding.  
At the start of each trial, we randomly sample the target index for the singulation target. We thus desire \emph{one} policy $\pi_\theta$ that can singulate \emph{any} of the blocks in $\mathcal{B}$.

\vspace{-3pt}
\section{Method: Singulating Objects in Packed Environments}

Our proposed framework (Figure~\ref{fig:overview_of_system}) uses high-DOF manipulators which make frequent contact with the target and other objects. Thus, to avoid directly modeling the nonlinear and frequent contacts, we use model-free reinforcement learning. We use the 16-DOF Allegro Hand, though in principle, we could use our method with other hands with simulation support. 
Our novel procedure involves careful design of the state (Section~\ref{ssec:state}) and multi-phase RL (Section~\ref{ssec:multi_phase_RL}). 

\subsection{State Representation}
\label{ssec:state}

\begin{figure*}[t]
\centering
\includegraphics[width=1.00\textwidth]{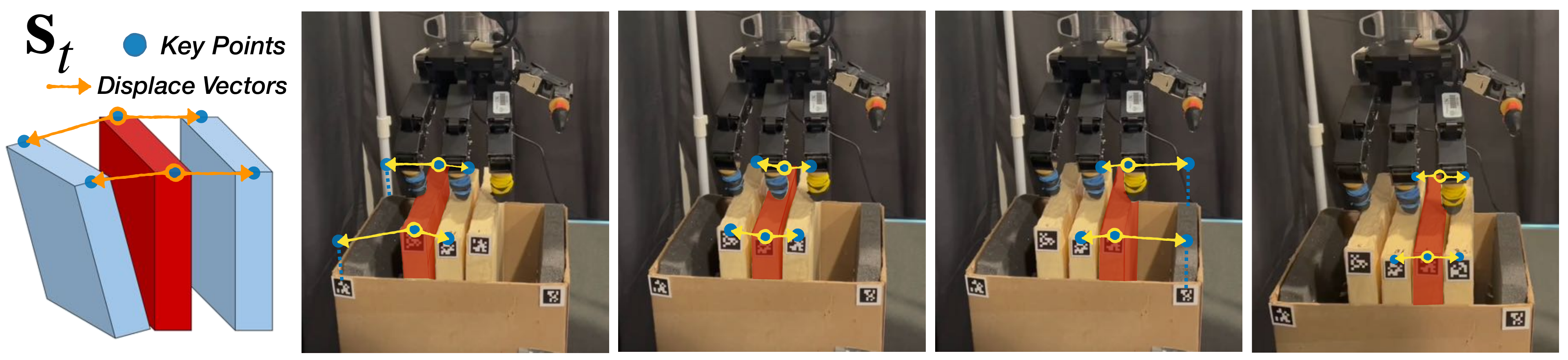}
\caption{
  Visualizations of the ``Block Information'' part of the state representation $\bs_t$. We show a clean view without the hand (left) followed by four examples of the state in physical experiments. The representation focuses on the target (colored red) and the displacement to its neighbors. By ``block corners,'' we refer to the top of the AprilTag markers, or virtual points above the box endpoints, as indicated with the arrows. %Furthermore, it also enables one policy that can singulate any of the possible blocks. %The figure is best viewed zoomed-in.
}
\vspace{-10pt}
\label{fig:state}
\end{figure*}

See Table~\ref{tab:observation} for the components of our state representation $\bs_t$, which we concatenate into one vector $\bs_t \in \mathbb{R}^{44}$. 
%This state conveys information about the environment to the robot's policy $\pi_\theta$. 
The ``Allegro Hand DOFs'' are the 16 real-time joint positions of the hand. 
The ``Block Information'' refers to the block corners and the displacement to neighboring block corners (Figure~\ref{fig:state}). 
%``Hand EE Information'' contains the real-time target top-edge position and the real-time Allegro Hand position calculated by adding a constant offset to the Franka end-effector position. 
``Hand EE Information'' contains the real-time target top-edge position and the real-time Allegro Hand pose, calculated by adding a constant offset to the standard reference point of the Franka EE. 
Finally, ``Policy Phase Indicator'' is an integer index of the policy's current phase. It also denotes whether the Allegro Hand is fixed, as the hand will be fixed during phase 3 (see Section~\ref{ssec:multi_phase_RL}).
We process the state by centering the observation around the initial ``top edge'' of the target block. This helps the policy be invariant to the initial block positions. 

% Daniel note: will have to adjust positioning of this table to make the LaTeX look nicer.
\begin{wraptable}{r}{0.46\textwidth}
%\begin{table*}[t]
\vspace*{-20pt}
  \setlength\tabcolsep{5.0pt}
  \centering
  \footnotesize
  \begin{tabular}{cc}
    \toprule
    Indices & Observation Description \\ 
    \midrule
    $[\;0, \,16)$   &  Allegro Hand DOFs \\
    $[16, 34)$  &  Block Information \\
    $[34, 43)$  &  Hand EE Information \\
    $[43, 44)$  &  Policy Phase Indicator \\
    \bottomrule
  \end{tabular}
  \caption{
  State representation $\bs_t$ for RL. 
  }
\vspace{-20pt}
\label{tab:observation}
%\end{table*}
\end{wraptable}
For the ``Block Information'' part of $\bs_t$, we consider keypoints of the target block and those directly adjacent to it. 
Concretely, for each block $b_i$, we consider its keypoints to be the position of its two topmost corners (indicated with circles in Figure~\ref{fig:state}). We can obtain this using ground-truth state information in simulation or by using object detection in the real world (see Section~\ref{ssec:sim2real}). %our implementation choice is to use AprilTag markers~\cite{olson2011tags}. 
In addition, our state representation includes information from \emph{adjacent} blocks $b_{i-1}$ and $b_{i+1}$, or the fixed walls of the container, if $i=1$ or $i=n$. Instead of directly using their keypoints, we use \emph{3D vectors} that ``point'' from keypoints of $b_k$ to the corresponding keypoints of $b_{k-1}$ and $b_{k+1}$ (see the arrows in Figure~\ref{fig:state}). 

Intuitively, $\bs_t$ is an efficient state representation that \emph{focuses on the target object} and \emph{its displacement to adjacent objects}. 
This facilitates simulation-to-real (sim2real) transfer as compared to using an image representation, due to the visual difference between images in simulation versus real.
Furthermore, this representation is scalable if we change either the number of total blocks or the target index $i$.  This representation is inspired by the concept of ``flow'' which describes how points move in space. Prior work has leveraged flow for manipulation of articulated objects~\cite{EisnerZhang2022FLOW} and tools~\cite{Seita2022toolflownet}; other works have used flow to model movement of objects in video frames~\cite{wen2024anypoint,yuan2024general}. 
In contrast, we do not train a model to predict object movement. We use flow in the reward function to indicate the \emph{separation} of a target object and adjacent objects, which also contrasts with~\cite{zhou2023hacman} who use flow to reward moving an object to a goal pose.

\subsection{Multi-Phase Design and Reward for Policy Training}
\label{ssec:multi_phase_RL}

To develop the policy, we propose to divide the task into three phases. In \emph{phase 1}, the hand needs to isolate the target block from its neighboring blocks. In \emph{phase 2}, the hand needs to tightly grasp the block. Finally, in \emph{phase 3}, the arm lifts the block. We train all three phases using reinforcement learning in simulation. 
All phases are the same except that they use different reward functions and Franka end-effector positions. During phase 3, the output of the policy is masked out, preventing any action from being applied to the Allegro Hand. Thus, the Allegro Hand's pose is fixed (while the arm moves).
The phases proceed sequentially, \ie the robot starts at phase 1, then moves to phase 2, then to phase 3.

We will leverage an off-the-shelf reinforcement learning algorithm, Proximal Policy Optimization (PPO)~\cite{ppo}. 
We define our reward $r_t$ at time $t$ as: 
\begin{equation}\label{eq:reward}
    r_t = w_h r_h + w_p r_p + w_g r_g + w_s p_s + w_o p_o + w_a p_a,
\end{equation}
which is the (phase-dependent) weighted sum of the six reward components: 
%The ``$w$'' terms are phase-dependent weights. 
\begin{align}
    r_h &= 
    \begin{cases}
    \min\left(h_{\text{target}}, \alpha_h\right), & \text{if } h_{\text{target}} \geq 0 \\
    \lambda_h * h_{\text{target}}, & \text{otherwise}
    \end{cases} & \text{Height Reward} \\
    r_p &= \exp\left(-\lambda_p * \max\left(\sum_{\,i=0}^{4} d_i, \alpha_p\right)\right) & \text{Proximity Reward} \\
    r_g &= \min\left(\sum_{\,i=0}^{4} c_i, \alpha_c\right) & \text{Grasp Reward} \\
    p_s &= \max\left(\alpha_s - d_s, 0\right) & \text{Split Penalty} \\
    p_o &= \sum_{\,i = 0}^{n} (h_i \, | \, i \neq i_{\text{target}} \, , \, h_i > \alpha_o) & \text{Other Blocks Penalty} \\
    p_a &= \left\| \mathbf{a} \right\|_2^2 & \text{Action Penalty}
\end{align}

\noindent with definitions and values in Table~\ref{tab:robot_parameters}. 

\begin{table}
\centering
%\small
\footnotesize
\begin{tabular}{>{\centering\arraybackslash}p{1.5cm} p{9cm} >{\centering\arraybackslash}p{1.5cm}}
\toprule
Symbol & \multicolumn{1}{c}{Description} & Value \\
\midrule
$h_{\text{target}}$ & Height of the target block & - \\
$\alpha_h$ & Height threshold & 0.2 \\
$\lambda_h$ & Scaling factor for negative height penalty & 0.1 \\
$\lambda_p$ & Scaling factor for proximity reward & 15 \\
$\sum_{i=0}^4 d_i$ & Distance of the fingertips from the center of the target block & - \\
$\alpha_p$ & Proximity threshold & 0.07 \\
$\sum_{i=0}^4 c_i$ & Number of fingertips in contact with the block & - \\
$\alpha_c$ & Contact threshold & 2 \\
$d_s$ & Minimum distance between any target and neighbor corner & - \\
$\alpha_s$ & Split distance threshold & 0.04 \\
$\alpha_o$ & Neighboring block distance threshold & 0.05 \\
%$n$ & Number of blocks & - \\
%$\mathbf{a}$ & Action taken by the robot & - \\
\bottomrule
\end{tabular}
\caption{Variables and parameters for our reward function (Equation~\ref{eq:reward}).}
\label{tab:robot_parameters}
\vspace{-25pt}
\end{table}

\vspace{-5pt}
\paragraph{Reward Rationale and Details.}

%\hao{TODO: Check and merge the rationale for the reward design}

The \emph{height} reward provides a positive reward if the target block is higher. %while penalizing it for a lower height. % (dropping) if $h_{\text{target}}$ is positive, we provide a positive reward capped at $\alpha_h$ (0.2); if $h_{\text{target}}$ is negative, we apply a penalty scaled by $\lambda_h$ (0.1). This provides a high reward if the robot lifts the target bloc, and penalizes it for dropping. %way, we can provide large reward if the robot lift the block and penalizes slightly if the block drops.
The \emph{proximity} reward incentivizes the fingertips to remain close to the target block by exponentially increasing the reward as the combined distance $\sum_{i=0}^4 d_i$ decreases. %The scaling factor $\lambda_p$ (15) controls how fast the reward increases. The threshold $\alpha_p$ (0.07) ensures a minimum distance value, promoting stability in the learning process.
The \emph{grasp} reward (inspired by~\cite{DexPoint2022}) gives positive reward as the number of fingertips $c_i$ in contact with the block increases, which can promote more secure grasping. We set $\alpha_c=2$ to encourage at least two fingers to be in contact. %This design can promotes secure grasping of the target block.
The \emph{split} penalty (with a negative weight $w_s$) encourages separation of the target block from its neighbors by at least $\alpha_s$ (we set \SI{4}{\centi\meter}).
%we apply the penalty based on the minimum displacement vector mentioned in Section~\ref{ssec:state}. We set the split distance threshold $\alpha_s = 0.04$ to encourage separation of the target block from its neighbors (by at least \SI{4}{\centi\meter}). 
The \emph{other block} penalty discourages lifting non-target blocks. % , it accumulates for any block $h_i$ other than the target $t$ that exceeds a height threshold $\alpha_o$ (0.05). We discourages lifting blocks that are not the target. 
Finally, we use an \emph{action} penalty to make the hand movement more efficient. 

%In all phases, we set the proximity reward $w_p = 0.2$ to encourage the fingertips to interact with the target block. We set the other block penalty $w_o = -11$ to discourage lifting irrelevant blocks.
All phases have the proximity weight $w_p = 0.2$ and other block penalty weight $w_o = -11$. 
In phase 1, we set $w_s = -10$ to prioritize isolating. We also set $w_h = 0$ and $w_g = 0$ to avoid grasping and lifting.
In phase 2, we set $w_g = 0.1$ and $w_s = -5$ to prioritize grasping while keeping the target block separated. We also set $w_h = 0$ to avoid lifting.
In phase 3, we set $w_h = 10$ to provide a large height reward if lifting is successful.
We include the action penalty $w_a = -0.001$ only in phases 1 and 2 when the hand is moving, and set it to zero otherwise.

\vspace{-3pt}
\paragraph{Rationale for Multi-Phase Design.}
We use multiple phases since we have three discrete steps in singulation: isolating, grasping, and retrieving. 
To this end, we hard-code the Franka arm's motion by pre-computing three positions of the Franka end-effector (EE), \ie wrist, for each phase.  
The policy has information about the EE position relative to the target block. If the EE is at the initial position, the system is in phase 1; if the EE is lower relative to the initial position, the policy knows it is transitioning to phase 2. 
Despite no actions applied to the Allegro Hand, including phase 3 in RL training significantly boosts total rewards for successful lifting, thereby encouraging better actions in phases 1 and 2 for a firmer grasp. 
As we show in Section~\ref{ssec:physical_results}, this multi-phase design outperforms the alternative where we merge phases.

\subsection{Simulation and Simulation-to-Real (sim2real) Transfer}
\label{ssec:sim2real}

% Daniel: removed `DexTouch2024`
We use Isaac Gym simulation~\cite{makoviychuk2021isaac}, since it enables massively parallel, GPU-accelerated reinforcement learning. 
Prior work has also used Issac Gym to train policies controlling the Allegro Hand for tasks such as block stacking~\cite{chen2023sequential}, tossing~\cite{dynamic_handover_2023}, and rotating objects~\cite{lin2024twisting,qi2022hand,touch-dexterity,RobotSynesthesia2024}. 
We deploy the learned policies directly in the real world using a physical Franka Panda arm with an Allegro Hand. 
To minimize the sim2real gap in the state, we use AprilTag markers~\cite{olson2011tags} and attach them to the corners of the blocks and the box containing them. In practice, this leads to relatively noise-free state estimation. 
In addition, we employ domain randomization in simulation~\cite{domain_randomization}, using parameters derived from~\cite{chen2023sequential}.  We randomize the spacing between blocks, and the initial Allegro Hand position. 
%We also randomize which target block to singulate. 

For the real world, we 3D print fingertips following~\cite{touch-dexterity}. These are thinner than the default Allegro fingertip pads, which can facilitate singulation. 
%We only move the fingertips, and pre-program or fix the arm's movement. 
%Furthermore, we had to adjust the offset of the Allegro Hand from the Franka robot in simulation versus in real. %
We also tune the Proportional-Derivative (PD) parameters by recording and visualizing the joint position of the 16 Allegro Hand motors. We compare the sinusoidal curves in simulation versus real, and adjust the $K_p$ and $K_d$ accordingly to minimize this difference. 
To reduce jittering in the hand motion, we apply an exponential moving average (EMA) smoothing factor of 0.2 to the policy's output. 

\begin{comment}
See the appendix and supplementary material for more details on the simulation task, the RL training parameters, and the sim2real pipeline.
\end{comment}

\vspace{-5pt}
\paragraph{What Didn't Work.}
We do not train $\pi_\theta$ to control the hand \emph{and} the Franka arm. While this succeeded in Isaac Gym simulation, the sim2real gap was too extreme due to differences in control between the arm and the hand. We failed to get any success in the real world in our preliminary trials. 
We also tried to use tactile sensing as done in~\cite{touch-dexterity} with low-cost force-sensing resistors, but we did not observe a benefit in preliminary real world trials. This may be due to the locations where the fingers make contact with objects in our experiments.

\section{Experiments}

\begin{figure}[t]
\centering
\includegraphics[width=1.00\textwidth]{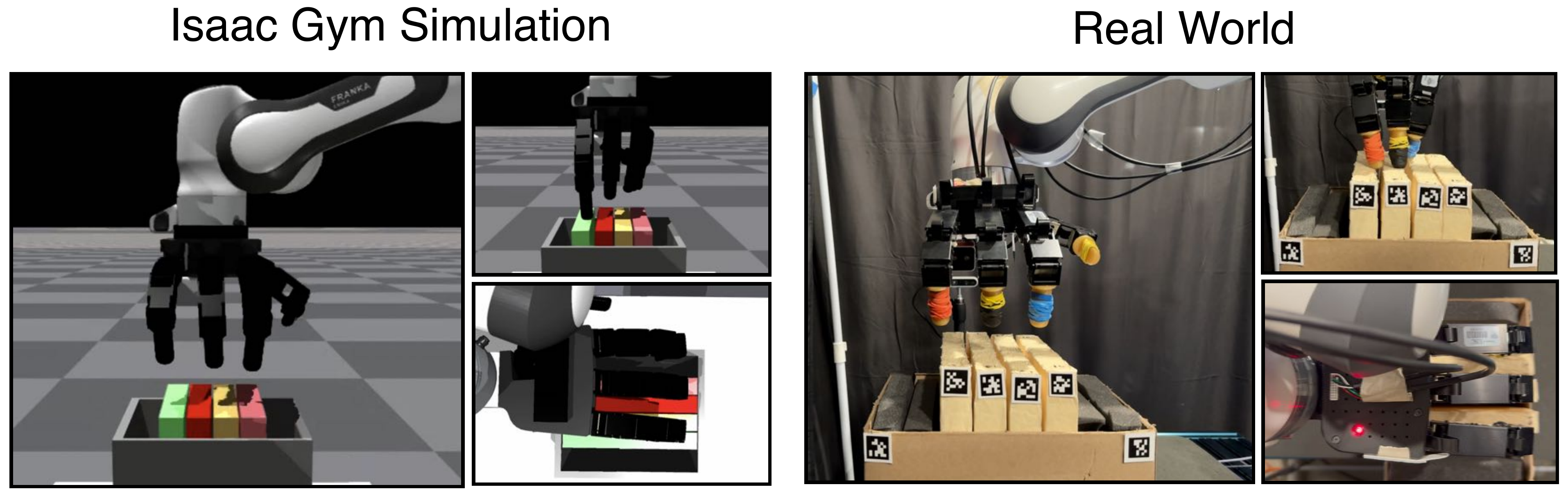}
\caption{
  Comparison of our simulation (left) and real world (right) setups. In Isaac Gym simulation~\cite{makoviychuk2021isaac}, we set up a robot with an Allegro Hand and create a box with blocks in it. In the real world, we attach a physical Allegro Hand to a Franka, and set up similar blocks within a cardboard box. We attach AprilTag markers to the blocks. % to facilitate position estimation for $\bs_t$.
}
\vspace{-13pt}
\label{fig:sim_vs_real}
\end{figure}

\subsection{Simulation Setup}

We implement our singulation task in Isaac Gym~\cite{makoviychuk2021isaac} by simulating a rigid container with rigid blocks (see Figure~\ref{fig:sim_vs_real}).  
We initialize the hand so that it starts above the target block. At the start of each episode, we sample the target block out of the set of blocks. %and provide the corresponding state representation to the policy. 
We set the simulation frequency at \SI{60}{\hertz} and the control frequency at \SI{20}{\hertz}.
As discussed in Section~\ref{ssec:multi_phase_RL}, we use a multi-phase learning procedure in simulation. 
Each episode (\ie trial) in simulation lasts 120 time steps. 
In simulation, we use the first 45 time steps for the phase 1 reward, the next 25 time steps for the phase 2 reward, and finally, the remaining time steps for the phase 3 reward.

\subsection{Simulation Experiments and Methods}
\label{ssec:sim_exp_methods}

We define a \textit{success} as keeping the target block \SI{20}{\centi\meter} above its initial position for 30 consecutive time steps, while other blocks remain below \SI{5}{\centi\meter}. 
%This design guarantees that one trial is a \textit{success} only if it isolates, grasps, and retrieves the target block successfully. 
%The \textit{success rate} is the number of successful singulation trials over total trials.
We evaluate multiple ablations in simulation to verify different parts of our pipeline. We compare our method with the following:

\begin{itemize}[label=$\bullet$]
    \item With Tactile Data: We integrate binary tactile sensor inputs into the observation space, as done in~\cite{touch-dexterity} except we add them to the four fingertips only. This changes $\bs_t$ from a vector in $\mathbb{R}^{44}$ to a vector in $\mathbb{R}^{48}$. %This results in 4 additional elements for our state representation.
    \item Naive Block Representation: We represent the blocks using the position and orientation (quaternion) of the target and neighboring blocks, instead of the top corners and displacements representation (see Section~\ref{ssec:state} and Figure~\ref{fig:state}).
    \item Two Phase Design: We combine phases 1 and 2 into a single phase, covering $45+25=70$ time steps, with the reward function from both phases also merged. Phase 3 remains unchanged. In this new two-phase approach, the first phase encourages both isolating and grasping actions, with the hand starting at the midpoint of the height previously used in the original Phases 1 and 2. The second phase is identical to the original Phase 3.
\end{itemize}

We run five random seeds of RL training. We record the online success rate of the last 100 episodes for singulating target block out of four blocks. %The results are averaged on 5 seeds. See Figure~\ref{fig:results-sim} for the training curves. 
At test time, we deploy our policy on a set of different environments to evaluate the generalization capabilities of our model in simulation. We test singulating ``1-out-of-1,'' ``1-out-of-2,'' ``1-out-of-5,'' and ``1-out-of-10'' tasks in a zero-shot manner and record the total success rate among 100 episodes.

\subsection{Simulation Results}

\begin{figure}[t]
\centering
\includegraphics[width=0.75\textwidth]{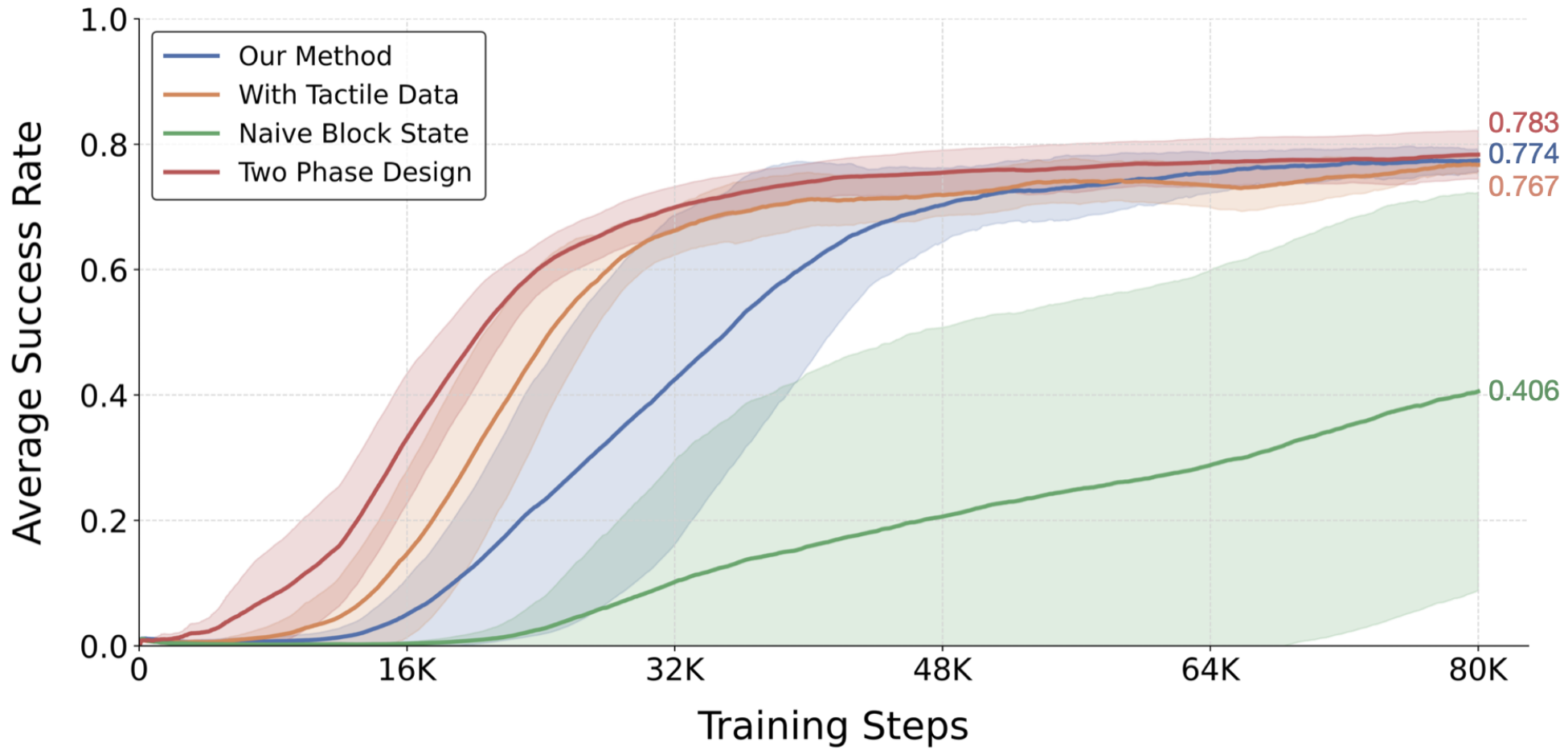}
\vspace{-10pt}
\caption{
  Policy performance curves for our method and alternatives; see Section~\ref{ssec:sim_exp_methods} for more details. Each curve shows the average online success rate over the last 100 episodes. Results are averaged over 5 seeds, and shaded areas show standard deviations. %each with 8192 parallel environments.
}
\vspace{-15pt}
\label{fig:results-sim}
\end{figure}

\begin{table*}[t]
  \setlength\tabcolsep{5.0pt}
  \centering
  \footnotesize
  \begin{tabular}{lcccc}
    \toprule
    Method & 1-out-of-1 & 1-out-of-2 & 1-out-of-5 & 1-out-of-10 \\ 
    \midrule
    With Tactile Data  & $0.84 \pm 0.02$ & $0.84 \pm 0.02$ & $0.81 \pm 0.02$ & $0.78 \pm 0.03$ \\
    Naive Block State  & $0.58 \pm 0.25$ & $0.57 \pm 0.27$ & $0.50 \pm 0.27$ & $0.45 \pm 0.26$ \\
    Two Phase Design   & $0.84 \pm 0.02$ & $0.86 \pm 0.03$ & $0.81 \pm 0.03$ & $0.79 \pm 0.03$ \\
    Ours               & $0.86 \pm 0.03$ & $0.86 \pm 0.02$ & $0.82 \pm 0.02$ & $0.78 \pm 0.01$ \\
    \bottomrule
  \end{tabular}
  \caption{
    Average success rates with standard deviations for different methods (left column, see Section~\ref{ssec:sim_exp_methods}) and different tasks at test time. Results average over 5 seeds.
  }
  \label{tab:sim_results}
  \vspace{-15pt}
\end{table*}

See Figure~\ref{fig:results-sim} for learning curves in simulation for different methods.
The results suggest that using binary tactile data does not offer improvements. %so we decide to not use tactile input in the real world to avoid potential sim2real gap.
We also observe that our corner-displacement state representation for the blocks leads to more sample-efficient training than our naive pose-based representation. 
The two phase design has a higher success rate in simulation compared to our method (0.783 versus 0.774, respectively). Interestingly, when we deploy this design in the real world (see Section~\ref{ssec:physical_results}), the success rate substantially drops, suggesting that having separate phases for isolating and grasping is better for sim2real.

See Table~\ref{tab:sim_results} for the success rates in singulating one block among different total blocks. As expected, the success rate decreases as the task complexity increases with more blocks. Nevertheless, the performance remains satisfactory even when singulating a target block out of a set of 10 blocks. This suggests that our policy effectively generalizes across varying scenarios. This outcome is also consistent with the design of our state representation, which remains invariant to the total number of blocks and focuses primarily on the target and its neighboring blocks.

%We encourage the reader to view the supplementary project website for more results and videos. % <-- Daniel: we already say this later and it's more important to say this during the physical experiments.

\subsection{Physical Setup}

See Figure~\ref{fig:sim_vs_real} for our real-world experiment setup. We consider a flat workspace with length \SI{122.5}{\centi\meter} and width \SI{82.5}{\centi\meter}.
We use a cardboard box with height \SI{17.5}{\centi\meter}, length \SI{26}{\centi\meter}, and width \SI{20}{\centi\meter}. 
For blocks, we cut out Styrofoam pieces which are \SI{3}{\centi\meter} thick. We also apply tape over the blocks to reduce the chances of them getting damaged by the robot hand.
To simplify experimentation, we fix the cardboard box on the workspace.  
We use a Panda Franka arm with an Allegro Hand. % (v4 edition). 
We mount two Intel RealSense D435 RGBD cameras at a height of \SI{35.5}{\centi\meter}, and to the sides at a distance of \SI{44}{\centi\meter} away from the nearest points of the cardboard box.  Finally, we apply AprilTag markers~\cite{olson2011tags} to the corners of the blocks and the sides of the cardboard box, to facilitate state estimation. 

\begin{figure}[t]
\centering
\includegraphics[width=1.0\textwidth]{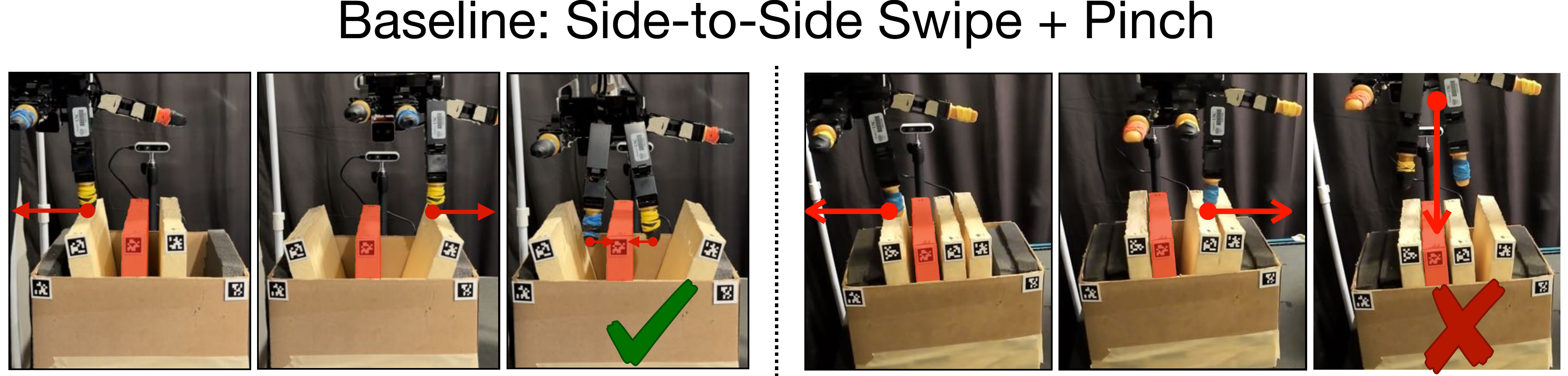} \caption{
  Two examples of frame-by-frame visualizations of the non-learning baseline S2SS\&P (target block colored red).  
  \textbf{Left}: S2SS\&P can succeed assuming sufficient space to push blocks. \textbf{Right}: S2SS\&P can struggle in more constrained setups. Pushing the two adjacent blocks results in limited changes compared to the original setup and the fingers would ``jam'' into the blocks during the attempted grasp. 
  %We call these experiment setups ``normal'' and ``constrained'', respectively, in Section~\ref{ssec:physical_methods}.
}
\vspace{-10pt}
\label{fig:physical-baselines}
\end{figure}

\subsection{Physical Experiments and Methods}
\label{ssec:physical_methods}

We perform the following physical experiments. 

\textbf{First}, we test our method and an ablation of it, which uses the two-phase reward design training in simulation. This is the zero-shot sim2real version of ``Two Phase Design'' in Section~\ref{ssec:sim_exp_methods} and reported in Table~\ref{tab:sim_results}. 
We consider singulating one block out of three, and do 15 singulation trials for each block (per method), resulting in $(15 \times 3) + (15 \times 3) = 90$ total trials.

\textbf{Second}, to investigate generalization capabilities, we test the \emph{same} policy from our method in singulating one item out of four blocks (instead of three). We do 10 singulation trials per block target, resulting in 40 trials. 
Furthermore, we compare our method with a baseline that does not use machine learning.  
%We first considered a baseline which detects the target block and extends the fingers of the hand to go ``straight down'' to grasp it. This reliably succeeds if there is a sufficiently large gap between the block and its neighbors. However, in preliminary experiments, we found that this method consistently failed if the blocks were too close to each other (as they are in our experiments), and we risked damage to the fingers as they ``jammed'' into the blocks. 
This baseline, \emph{``Side-to-Side Swipe and Pinch''} (S2SS\&P), uses one finger to push aside blocks next to the target to create a gap. 
See Figure~\ref{fig:physical-baselines} for the S2SS\&P baseline. 
This leads to $40+40=80$ trials. As with the first set of experiments, we conduct these 80 trials under a ``normal'' environment where there is sufficient space for the S2SS\&P baseline to reliably succeed; this environment looks like Figure~\ref{fig:physical-baselines} (left), but with four blocks. 
% during which the baseline reliably succeeds if there is enough clearance to push aside neighboring blocks (see Table~\ref{tab:physical_results_4}).
To challenge both methods, we consider a \emph{more constrained environment} where we apply extra foam inside the box (see Figure~\ref{fig:physical-baselines}, right). %to create a more constrained environment. 
We perform 40 trials (ten per block) for our method and S2SS\&P under this ``constrained'' environment, resulting in $40+40=80$ additional trials. 
% \hao{We should further explain what 'Normal' and 'Constrained' environment means. Maybe we should also add explanation in caption of each table}

\begin{table*}[t]
  \setlength\tabcolsep{5.0pt}
  \centering
  \footnotesize
  \begin{tabular}{lccccc}
    \toprule
    Environment & Method & Front & Middle & Back & Overall \\ 
    \midrule
    Normal & Two Phase Design & 2/15 & 3/15 & 5/15 & 10/45 \\
    Normal & Ours             & 13/15 & 14/15 & 13/15 & 40/45 \\
    \bottomrule
  \end{tabular}
  \caption{
    Performance of our method and a baseline in physical experiments, when singulating one block out of three. 
    We conduct 45 trials per method, with 15 each for ``Front,'' ``Middle,'' and ``Back'' blocks. We obtain $40/45\approx 0.888$ success compared to $10/45\approx 0.222$ for the baseline. 
    See Sections~\ref{ssec:physical_methods},~\ref{ssec:physical_results}, and~\ref{ssec:failures} for more details. 
  }
  \label{tab:physical_results_3}
  \vspace{-10pt}
\end{table*}

\begin{table*}[t]
  \setlength\tabcolsep{5.0pt}
  \centering
  \footnotesize
  \begin{tabular}{lcccccc}
    \toprule
    Environment & Method & Front & Middle-1 & Middle-2 & Back & Overall \\ 
    \midrule
    Normal & S2SS\&P & 10/10 & 6/10 & 10/10 & 10/10 & 36/40 \\
    Normal & Ours & 9/10 & 8/10 & 9/10 & 7/10 & 31/40 \\
    Constrained & S2SS\&P & 0/10 & 0/10 & 0/10 & 0/10 & 0/40 \\
    Constrained & Ours & 7/10 & 6/10 & 5/10 & 10/10 & 28/40 \\
    \bottomrule
  \end{tabular}
  \caption{
    Performance of our method compared with the S2SS\&P baseline in physical experiments, when singulating one block out of four under a ``normal'' (top two rows) or ``constrained'' (bottom two rows) environment. Our method uses the \emph{same} policy for singulating one out of three, as reported in Table~\ref{tab:physical_results_3}. We report the middle blocks separately as ``Middle-1'' and ``Middle-2.'' 
    See Sections~\ref{ssec:physical_methods},~\ref{ssec:physical_results}, and~\ref{ssec:failures} for more details. 
    %Of the 9 failures, 8 were failures from ``separating'' while the last was due to a failure during the ``grasping \& lifting'' portion after successful separating. 
  }
  \label{tab:physical_results_4}
  \vspace{-20pt}
\end{table*}

% Daniel: experiment counts
% First: 45 + 45 = 90
% Second: 40 + 40 + 40 + 40 = 160
% Total: 90 + 160 = 250.
All together, these experiments contain \textbf{250 physical trials}. 
For each trial, we manually evaluate whether the robot was able to lift the target block entirely above all other blocks. We categorize failures into ``isolating'' or ``grasping and retrieving'' failures. 
To set a new bar for experiment transparency, we report our full experiment record (with videos of all trials) on the project website.

\subsection{Physical Results}
\label{ssec:physical_results}

See Table~\ref{tab:physical_results_3} for quantitative results from our first set of experiments. 
Our method obtains a success rate of $40/45\approx 0.889$, of which two failures were from the \emph{isolating} step, and three were from \emph{grasping and lifting}.  
Breaking it down by the singulation target, our method attains roughly equal performance among the blocks, getting $13/15$, $14/15$, and $13/15$ for the front, middle, and back blocks. 
In contrast, the Two Phase Design baseline struggles significantly, attaining just a $10/45\approx 0.222$ success rate, despite how it outperformed our method in simulation. 
It has a $32/45 \approx 0.711$ success rate after the first stage, \emph{isolating}. However, it performs weaker grasps, causing more failures in \emph{grasping and lifting}, thus lowering the overall singulation success to $10/45$.  
This suggests that our multi-phase RL design encourages the robot to obtain a firmer grasp. 

% \begin{wraptable}{r}{0.30\textwidth}
% \vspace*{-20pt}
%   \setlength\tabcolsep{5.0pt}
%   \centering
%   \footnotesize
%   \begin{tabular}{cc}
%     \toprule
%     Method & Success \\ 
%     \midrule
%     S2SS\&P &  0/4 \\
%     Ours     &  4/4 \\
%     \bottomrule
%   \end{tabular}
%   \caption{
%     Our method versus a baseline. 
%   }
% \vspace{-20pt}
% \label{tab:baseline-compare}
% \end{wraptable}

We report results for our second set of experiments in Table~\ref{tab:physical_results_4}, where we evaluate our method in singulating one out of \emph{four} blocks. Using the \emph{same} policy as in Table~\ref{tab:physical_results_3}, our method obtains $31/40 = 0.775$ success rate, with roughly consistent performance across the four blocks. It performs worst with $7/10$ success on the ``back'' block. 
In addition, of the 9 failures, 8 occurred during \emph{grasping and lifting}. 
These results are promising, and suggest that our state representation helps the policy generalize to new scenarios with more blocks by capturing essential task information.  
While the success rate of $31/40$ is lower compared to $40/45$, this may be because using an extra block results in increased resistance. 

\begin{figure}[t]
\centering
\includegraphics[width=1.00\textwidth]{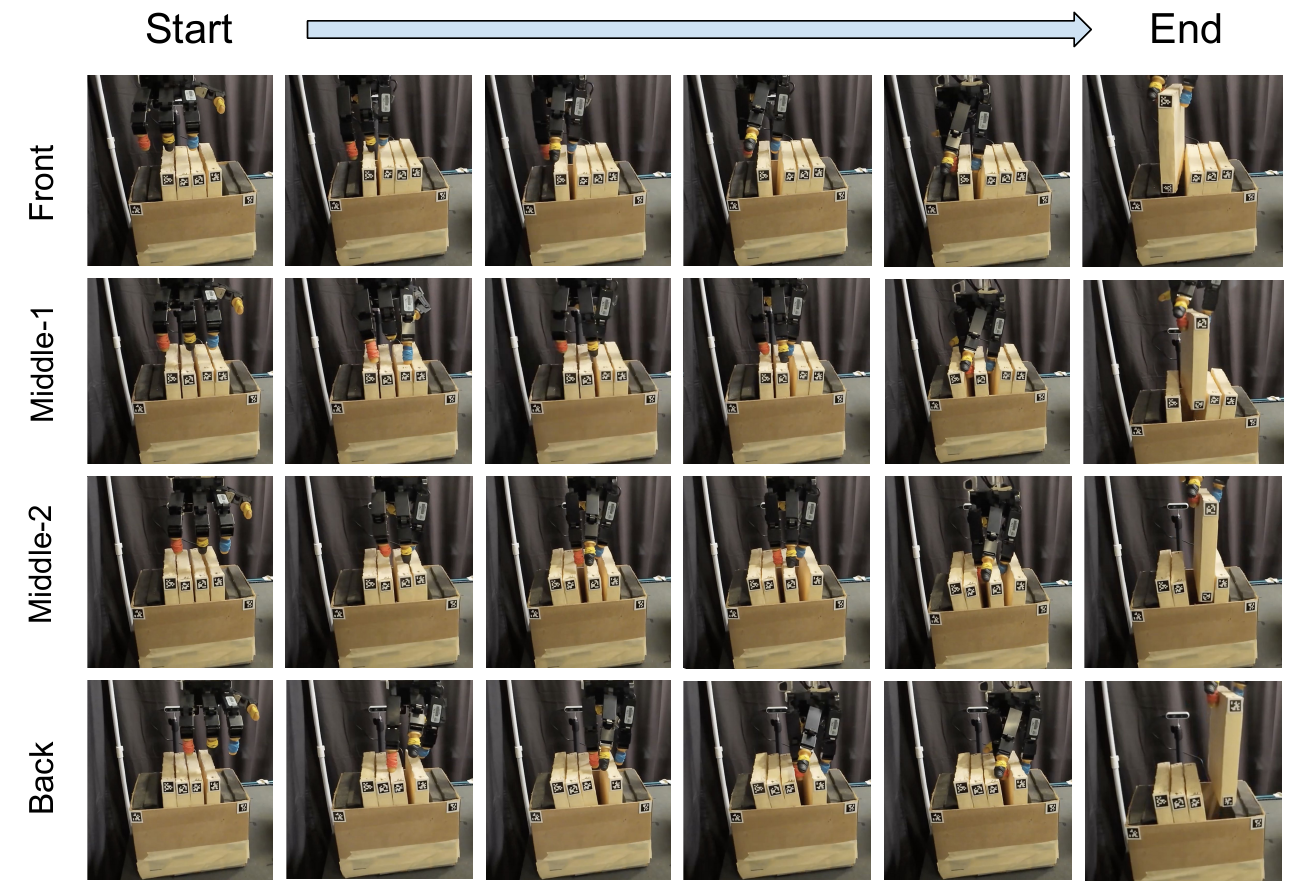}
\caption{
  Example successful singulation rollouts (one per row) of our system using the Allegro Hand. 
  We deploy one policy trained entirely in simulation to singulate the target. Each row shows a different target block (for the same policy). As viewed in the image, the blocks are (from left to right): Front, Middle-1, Middle-2, Back. This is the ``constrained'' environment as discussed in Section~\ref{ssec:physical_methods} with extra foam in the box.
}
\vspace{-10pt}
\label{fig:teaser}
\end{figure}

Table~\ref{tab:physical_results_4} also compares our method with the non-learning baseline (S2SS\&P), visualized in Figure~\ref{fig:physical-baselines}. 
Under normal conditions, the baseline method achieves a 36/40 success rate, whereas our method achieves a 31/40 success rate.
In the constrained environment, however, our method only has a minor drop (28/40 success), while S2SS\&P fails completely (0/40). 
The key advantage of our method is that it can result in fingers pushing adjacent blocks to temporarily create space, while simultaneously lowering the fingers. The baseline is unable to create enough clearance, as the blocks tend to reset back to their original position. 
See Figure~\ref{fig:teaser} for example rollouts from our method in this constrained environment. 

%As it is much easier to get intuition from looking at videos, 
We encourage the reader to consult the project website to view videos. 
%We also show some experiments where the policy is robust to a human which disturbs the scene.
%Finally, we notice a trend that the hand gets slightly progressively worse performance over time. For example, midway through some trials, we observed that a screw had come off from an earlier set of trials, and thus we had to discard this data and re-assemble the hand. 

\subsection{Failure Cases and Limitations}\label{ssec:failures}

\begin{figure}[t]
\centering
\includegraphics[width=1.00\textwidth]{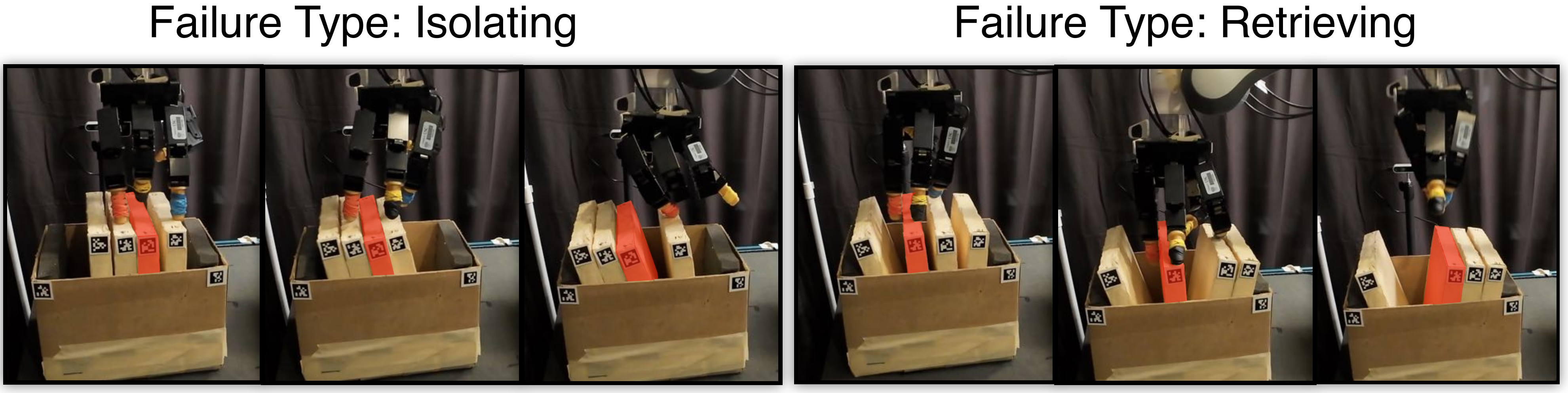}
\caption{
  Example failure cases of our method. \textbf{Left}: the target is the third block from left, but the policy experiences an isolating failure due to fingers getting stuck and an inability to ``recover.'' 
  \textbf{Right}: the target is the second block from left, but the policy fails to sufficiently grasp the block, and thus fails during the retrieving stage. 
}
\vspace{-10pt}
\label{fig:failure}
\end{figure}

As reported in Section~\ref{ssec:physical_results}, our method can sometimes fail. 
For example, the Allegro Hand can get stuck if it reaches an out-of-distribution state in the real world. 
The fingers can also get stuck when pushing against the blocks. 
We visualize two representative failures of our method in Figure~\ref{fig:failure}.  
Our method also has several limitations, which present exciting opportunities for future work. 
First, even though we reduce the sim2real gap by using keypoints instead of images, we still rely on considerable real world engineering of the physical setup 
%(\eg careful placement of blocks and the camera) 
to enable sim2real with keypoints while avoiding occlusions. 
Second, our representation assumes that we can specify an object's state with keypoints, but this may be challenging for other objects, especially deformable objects. 
Finally, our system assumes that it is fine to push aside the non-target blocks, but this assumption may not be valid when the adjacent obstacles are fragile (\eg glass vases). 
%We are bound by the limits of the Allegro Hand which may have difficulty with much finer-grained manipulation, as well as the current Isaac Gym simulation inaccuracies which for example cannot accurately simulate deformable objects and hence it would likely fail if we tried to singulate towels. 

\section{Conclusion}

We propose a framework, Singulating Objects in Packed Environments (\method), which involves isolating, grasping, and retrieving an item in clutter. 
Our pipeline consists of a displacement-based state representation with a multi-phase reward function, which together can facilitate object singulation using a dexterous high-DOF Allegro Hand. %We also use multiple phases with different reward functions. 
We develop a simulation environment in Isaac Gym and train a singulation policy using reinforcement learning. 
We show successful zero-shot simulation-to-real transfer, suggesting that our method can singulate items in different locations, as well as singulate a target among different amounts of nearby items. 
We hope this work inspires future research in dexterous manipulation and object singulation.

\section*{Acknowledgments}
\vspace{-5pt}
{\footnotesize
We thank Chen Wang, Yuzhe Qin, Haozhi Qi, and Binghao Huang for helpful advice on using the Allegro Hand with Isaac Gym and on setting up the real world experiments.
We thank Vedant Raval and Harshitha Rajaprakash for helpful writing feedback. 
}
\vspace{-5pt}

% Make the biblatex respect margin
\appto{\bibsetup}{\sloppy}
{\small
\printbibliography
}

\end{document}